\documentclass[runningheads,a4paper]{llncs}

\usepackage{times}
\usepackage{graphicx}
\usepackage{url}
\usepackage{cite}
\usepackage{amsmath,amssymb,amsfonts}
\usepackage{algorithmic}
\usepackage{textcomp}
\usepackage{xcolor}
\usepackage{booktabs}
\usepackage{multirow}
\usepackage{rotating}
\usepackage{array}
\usepackage{float}
\usepackage[ruled]{algorithm2e}
\usepackage[super]{nth}
\DeclareMathOperator*{\argmax}{argmax}

\newcommand{\keywords}[1]{\par\addvspace\baselineskip
\noindent\keywordname\enspace\ignorespaces#1}

\urldef{\mailsa}\path|{iuliia.nigmatulina, evillatoro}@idiap.ch|

\begin{document}

\title{Unifying Global and Near-Context Biasing in a\\
Single Trie Pass\thanks{This work was supported by Idiap Research Institute and
Uniphore collaboration project. Part of this work was also supported by EU Horizon 2020 project ELOQUENCE (grant number 101070558).}}

\titlerunning{Unifying Global and Near-Context Biasing}

\author{Iuliia Thorbecke$^{1, 2}$ \and Esa\'{u} Villatoro-Tello$^{1}$ \and Juan Pablo Zuluaga$^{1}$ \and Shashi Kumar$^{1, 3}$ \and Sergio Burdisso$^{1}$ \and Pradeep Rangappa$^{1}$ \and Andr\'{e}s Carofilis$^{1}$ \and Srikanth Madikeri$^{2}$ \and Petr Motlicek$^{1, 4}$ \and Karthik Pandia$^{5}$ \and Kadri Hacio\u{g}lu$^{5}$ \and Andreas Stolcke$^{5}$}


\authorrunning{Thorbecke, Iuliia, Villatoro-Tello, Esa\'{u}, \textit{et al.}}

\institute{Idiap Research Institute, Switzerland \\
\mailsa\\
\and
University of Zurich, Switzerland \\
\and
EPFL, Switzerland \\
\and
Brno University of Technology, Czech Republic \\
\and
Uniphore, India/USA \\
}

\index{Thorbecke, Iuliia}
\index{Villatoro-Tello, Esa\'{u}}
\index{Zuluaga, Juan Pablo}
\index{Kumar, Shashi}
\index{Burdisso, Sergio}
\index{Rangappa, Pradeep}
\index{Carofilis, Andr\'{e}s}
\index{Madikeri, Srikanth}
\index{Motlicek, Petr}
\index{Pandia, Karthik}
\index{Hacio\u{g}lu, Kadri}
\index{Stolcke, Andreas}

\toctitle{} \tocauthor{}

\maketitle

%
%
%
%
\begin{abstract}
Despite the success of end-to-end automatic speech recognition (ASR) models, challenges persist in recognizing rare, out-of-vocabulary words —including named entities (NE)— and in adapting to new domains using only text data.
This work presents a practical approach to address these challenges through an unexplored combination of an NE bias list and a word-level n-gram language model (LM).
This solution balances simplicity and effectiveness, improving entities' recognition while maintaining or even enhancing overall ASR performance.
We efficiently integrate this enriched biasing method into a transducer-based ASR system, enabling context adaptation with almost no computational overhead.
We present our results on three datasets spanning four languages and compare them to state-of-the-art biasing strategies.
We demonstrate that the proposed combination of keyword biasing and n-gram LM improves entity recognition by up to 32\% relative and reduces overall WER by up to a 12\% relative.

\keywords{Contextualisation and adaptation of ASR, real-time ASR, Aho-Corasick algorithm, Transformer-Transducer}
\end{abstract}

\section{Introduction}
Traditional methods for incorporating contextual text data can be broadly categorized into two approaches: (1) integrating the information during decoding without modifying the ASR architecture, such as \textit{rescoring} and \textit{shallow fusion} (SF)~\cite{aleksic2015bringing,hori2017multi,kannan2018analysis,zhao2019shallow,tian2022improving,jung2022spell,guo2023improved, wang2023contextual,10301513}, and (2) training the model to accept contextual input dynamically when needed~\cite{pundak2018deep,jain2020contextual,qiu2023context}.

We focus on the first group of methods, which can be applied to any ASR model and are considered the least costly methods of context integration.

SF means log-linear interpolation of the score from the E2E model with an external contextual LM at each step of the beam search:
\begin{equation}
  y^{\ast} = \argmax \log P(y|x) + \lambda \log P_C(y),
\end{equation}
where $P(y|x)$ is the probability of sequence $y$ predicted by the E2E model based on the input acoustic feature $x$. $P_C(y)$ is an in-domain or context-biased \textit{contextual} LM and $\lambda$ is a hyperparameter to control the impact of the contextual LM on the overall model score \cite{kannan2018analysis, zhao2019shallow}.
For the speed and convenience of decoding, contextual information is usually presented as a graph, such as an n-gram LM as a weighted finite-state transducer (WFST) \cite{mohri2002weighted,aleksic2015bringing,nigmatulina2023implementing} or
a bias list as a prefix trie \cite{jung2022spell}, and a string-matching algorithm is usually used to detect relevant context entities in the hypotheses.\footnote{https://github.com/kensho-technologies/pyctcdecode}

\begin{figure}[htbp]
    \centering
    \includegraphics[width=0.8\linewidth]{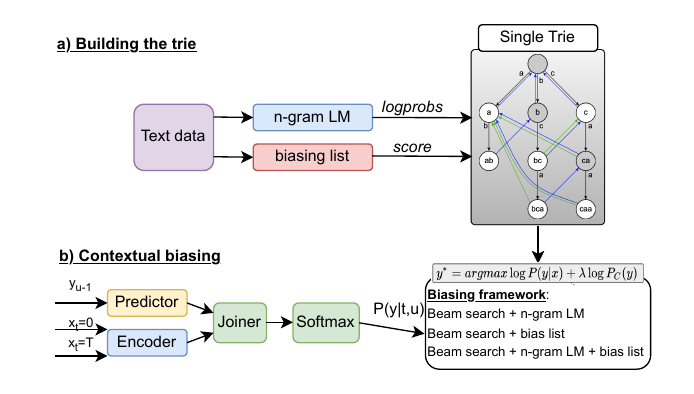}
    \caption{Proposed biasing approaches at beam search time with Aho-Corasick string matching algorithm, including biasing list and n-gram LM statistics. 
    }
    \label{fig:biasing-systems}
\end{figure}

In previous studies on SF, typically either the effect of an external LM or a keyword bias list is investigated.
In \cite{kannan2018analysis}, subword-level and grapheme-level neural network language models (NNLMs) are used for rescoring. In \cite{hori2017multi}, the authors combine the statistics over characters and words by rescoring with character- and word-level recurrent NNLMs  (RNNLMs) at the same time. Both approaches, however, are unsuitable for online recognition and fast context adaptation due to the size of the NNLMs. In \cite{tian2022improving}, the n-gram word-level LM is adopted by converting subwords to words before rescoring; this allows on-the-fly decoding but, as a character-level sequence can be mapped to different word sequences, during the decoding, the authors first have to convert character-level partial hypothesis into a word lattice.

On the other hand, SF with a list of target entities \cite{le2021deep,jung2022spell} can bias hypotheses towards particular words or NEs, such as proper names, terminology, or geographical names.
Context keywords can be integrated directly with the \textit{keyword prefix trie} search algorithm without a need to train any LM~\cite{jung2022spell}. Keyword biasing (KB), however, can lead to a degradation of the overall performance. 
Zhao~et~al.~\cite{zhao2019shallow} improve SF with biasing at the subword unit level instead of word level and use a set of common prefixes (\textit{``call''}, \textit{``text''}) to avoid irrelevant bias.

In our approach, we enhance KB while maintaining robust overall performance by leveraging \textit{``near-context''} n-grams from an n-gram LM alongside the \textit{``global-context''} entities.
Concretely, we unify existing SF methods by integrating a word-level language model with the SF of context entities within a \textit{single trie pass}. First, we construct a lightweight n-gram LM by dynamically converting words into subwords \textit{on-the-fly} and adjusting LM weights to correspond to the new subword units. Next, keyword weights are modified within the same trie, allowing for seamless integration of both LM n-grams and KB.  

This approach extends the KB method from \cite{guo2023improved} by merging keyword information with LM n-grams to form a unified context graph (Fig.~\ref{fig:biasing-systems}). We employ the Aho-Corasick (AC) algorithm for string matching, which efficiently handles search failures in the trie by managing mismatched cases. Given its proven efficiency in NLP tasks \cite{svete2023algorithms} and prior success in ASR KB \cite{guo2023improved}, we demonstrate its effectiveness for SF as well, ensuring robust and efficient context integration during decoding.



In sum, our main contributions are as follows: \textit{(i)} efficient contextual adaptation using a word-level n-gram LM and KB within a \textit{single context trie};  \textit{(ii)} comprehensive analysis of performance across named entities (NEs), out-of-vocabulary (OOV) NEs, and real-time factor (RTF) metrics; (iii) extending the use of the AC algorithm to integrate word-level n-gram LMs, previously applied only for KB; and \textit{(iv)} benchmarking multiple SF strategies across four languages and three datasets.

\section{Methodology} 
\label{sec:method}
\subsection{Preliminaries: The Aho-Corasick Algorithm}
The AC algorithm proposed by \cite{aho1975efficient} is a text pattern-matching algorithm that operates in linear time. It utilizes three data structures to represent the search set, or \textit{transition diagram}: a trie, an output table, and a failure function. The output table stores suffixes reachable from any node that corresponds to the target search strings. The failure function handles cases where some search strings are suffixes of others \cite{meyer1985incremental}. For example, if a trie contains the word ``\texttt{CAN}'' but not \texttt{CAT}'', a failure transition will backtrack ``\texttt{CAT}'' to the prefix ``\texttt{\textbf{CA-}}'' from ``\texttt{CAN}'' upon a mismatch.
The algorithm is implemented as a finite state machine with backoff arcs that not only return to the root or a lower-order n-gram (e.g., from a 4-gram to a 3-gram) when a string end is reached, but also perform \textit{failure transitions} to backtrack upon a mismatch. This mechanism allows partial match detection while keeping the trie’s structure sparse, improving efficiency by avoiding repeated prefix matches.\footnote{Our work is based on the existing Aho-Corasick implementation available in the k2/icefall framework (\url{https://github.com/k2-fsa/icefall/blob/master/icefall/context_graph.py}).}


\subsection{Word-level n-gram LM for Transducer-based ASR}
\label{subsec:method-ngram}

One of the key contributions of our approach is the integration of a \textit{word-level} n-gram language model, henceforth referred to as WL-3gramLM, into the Transducer architecture, a non-straightforward task given that the model operates at the BPE unit level \cite{sennrich2016neural}. Unlike traditional SF methods that rely on subword-level LMs, our proposed solution enables direct incorporation of a word-level n-gram LM, leveraging its richer word-level statistics and improved transparency for adaptation and biasing.

To achieve this, we propose an enhanced SF approach that utilizes the AC algorithm to construct a prefix trie directly from LM n-grams and their corresponding log probabilities. Since the ASR model produces hypotheses at the BPE level, we first convert the WL-3gramLM n-grams into sequences of BPE units using SentencePiece.\footnote{\url{https://github.com/google/sentencepiece}} During decoding, when a hypothesis string matches a stored n-gram in the context trie, we incorporate the LM probability score as an additional bias score into the log probability of the matched candidate. Algorithm~\ref{alg:context_scores} illustrate how the LM probability score (\texttt{bias\_score}) is obtained before the context trie is created. 


A key challenge in integrating an ARPA-formatted n-gram LM is determining how to convert its word-level scores into a bias score (bonus) that can be effectively combined with the ASR model's subword-level log-probability scores. An n-gram LM in ARPA format provides scores as log-base-10 probabilities, $s_w = \log_{10}(P_{LM}(w))$. Applying $s_w$ directly as a bias cost would lead to poorly scaled biasing. However, through empirical evaluation on our development sets, we found that a non-linear transformation of the LM score yielded superior performance.
We systematically compared several functions and discovered that the most effective bias score is derived by directly exponentiating the original log-base-10 score:

\begin{equation}
\label{eq:proposed_fusion}
\text{bias\_score}(w) = \exp(s_w) = \exp(\log_{10}(P_{LM}(w)))
\end{equation}


This transformation is mathematically equivalent to $P_{LM}(w)^{\frac{1}{\ln(10)}}$, which is approximately $P_{LM}(w)^{0.434}$. This effectively compresses the dynamic range of the original LM probabilities, emphasizing higher probabilities less aggressively than a direct linear scaling, and sharply diminishing the influence of less probable ones. While the exponent $1/\ln(10)$ arises naturally from converting a log-base-10 probability into the argument of the natural exponential function, we acknowledge that this implicit compression factor could be considered an arbitrary choice and, in a more generalized framework, might be treated as a tunable hyperparameter $\gamma$ (e.g., $P_{LM}(w)^\gamma$). However, for the scope of this work, this specific transformation proved robust and effective across our datasets.

Crucially, the $\text{bias\_score}(w)$ obtained through this transformation is a value in \textit{probability space}, not log-probability space. When this score is applied, it is added to the log-probability of the ASR model's hypothesis. Specifically, we treat this $\text{bias\_score}(w)$ as a positive bonus that directly increases the overall log-likelihood of the matched word sequence during beam search. This unconventional direct addition of a probability-space value to a log-probability serves as an aggressive non-linear scaling, heavily rewarding high-probability n-grams while allowing for more nuanced integration than a simple linear addition in the log-domain.

A second implementation challenge is to apply this word-level $\text{bias\_score}(w)$, to a word $w$ that has been tokenized into a sequence of subwords $(b_1, b_2, \dots, b_k)$. We explored two strategies: (a) distributing the bonus evenly across all subwords (i.e., adding $\text{bias\_score}(w)/k$ at each subword step) and (b) applying the entire bonus at a single point. Our experiments showed that the latter approach was more effective. Specifically, the entire bonus is added to the hypothesis score upon the emission of the final subword, $b_k$, of the word $w$.

To further analyze the impact of word-level statistics on subword-level decoding, we evaluate the model’s ability to recognize OOV named entities (\S\ref{subsec:training}). The AC algorithm’s capability to efficiently locate partial matches proves especially beneficial for improving OOV word recognition, reinforcing the advantage of our word-level LM integration over conventional subword-only approaches.

\subsection{Unifying n-Gram LM and KB via AC}
In this section, we present the integration of a word-level n-gram LM with KB during the decoding process. Specifically, we introduce a novel implementation strategy using a \textit{single context trie} to combine LM n-grams with target keywords, while maintaining traversal at the BPE level. 

During KB in decoding, a fixed positive score is added to the log probability of a candidate in the hypothesis when it matches any of the context entities. To facilitate this, we employ a trie structure derived from the AC algorithm, which enables efficient lookup and biasing. This trie contains both LM n-grams and the KB information. 

The integration of keywords with LM n-grams is described in Algorithm~\ref{alg:context_scores}. First, all n-grams from the LM are inserted into the trie with their respective scores. Then, for each keyword, a \textit{bias score} is applied depending on whether the keyword’s n-gram is present in the LM (\textit{``inLM''}) or not (\textit{``outLM''}). If the keyword n-gram is already present in the LM, its bias score is adjusted based on its associated LM score, while for keywords absent in the LM, a fixed bias score is applied.

The bias score is tuned on the development set, with different values for when the keyword n-gram is in the LM versus when it is not:
\[
final\_bias\_score = 
\begin{cases}
  \alpha_{outLM} & \text{if NE is not in LM,} \\
  exp(LMs) + \alpha_{inLM} & \text{if NE is in LM,} \\
  exp(LMs) & \text{other words in LM.}
\end{cases}
\]
where $LMs$ is the LM score, and $\alpha$ represents the fixed bias score. We conducted a grid search to determine the optimal values of the $\alpha$ hyperparameters. We evaluated different biasing scores for both cases: $\alpha_{inLM} \in \{0.5, 1.0, 1.5, 2.0\}$ and $\alpha_{outLM} \in \{0.5, 1.0, 1.5, 2.0\}$. Our results consistently show that the combination of $\alpha_{inLM}=0.5$ and $\alpha_{outLM}=1.5$ achieves the best performance across all experimental settings.

\begin{algorithm}[t]
{\scriptsize
\DontPrintSemicolon
\SetKwProg{Fn}{Function}{:}{}
\Fn{$\texttt{context\_file}$, $\texttt{arpa\_file}$, $\texttt{in\_lm\_cost}$, $\texttt{not\_in\_lm\_cost}$}{
    Initialize empty list \texttt{contexts} \;
    Initialize empty dictionary \texttt{contexts\_scores} \;
    Initialize empty dictionary \texttt{keywords} \;

    \ForEach{line in \texttt{context\_file}}{
        \texttt{keywords[line.strip()]} $\gets$ ``not in the lm''\;
    }

    \ForEach{line in \texttt{arpa\_file}}{
            Split line into \texttt{[weight, ngram]}\;
            Append \texttt{ngram} to \texttt{contexts}\;
            \texttt{bias\_score} $\gets e^{\texttt{float(weight)}}$\;

            \If{\texttt{ngram} in \texttt{keywords}}{
                \texttt{keywords[ngram]} $\gets$ ``in the lm''\;
                \texttt{contexts\_scores[ngram\_ids]} $\gets$ \texttt{bias\_score} $+$ \texttt{in\_lm\_cost}\;
                }
            \Else{
                \texttt{contexts\_scores[ngram\_ids]} $\gets$ \texttt{bias\_score}\;
            }
        }

    \ForEach{(\texttt{entity}, \texttt{label}) in \texttt{keywords.items()}}{
        \If{\texttt{label} is ``not in the lm''}{
            Append \texttt{entity} to \texttt{contexts}\;
            \texttt{contexts\_scores[ngram\_ids]} $\gets$ \texttt{not\_in\_lm\_cost}\;
        }
    }

    \texttt{context\_trie} $\gets$ \texttt{ContextTrie(contexts\_scores)}\;
    \texttt{context\_trie.build(SentencePiece.encode(contexts))}\;
\caption{Compute Context Scores from ARPA and Keyword Files}
\label{alg:context_scores}
}
}
\end{algorithm}



\section{Experimental Setup}
\label{sec:experiments}

\subsection{Data}
\label{subsec:data}

\begin{table}[t]
 \scriptsize
  \centering
  \caption{Test sets statistics with context information. $^{\dagger}$Number of utterances with at least one named entity.}
  \resizebox{0.6\textwidth}{!}{
  \begin{tabular}{ lccccc }
    \toprule
    \multirow{3}{*}{\textbf{Test set}} & \multirow{3}{*}{\textbf{Size}} & \textbf{Duration} & \multicolumn{3}{c}{\textbf{Biasing entities}} \\
    \cmidrule(lr){4-6}
     & & \textbf{(hours)} & \textbf{unique} & \textbf{NE-utts}$^{\dagger}$ & \textbf{OOV} \\
    \midrule
    DefinedAI & 2K utt. & 6 & 367 & 486 & 16 \\
    Earnings21 & 18K utt. & 39 & 1013 & - & - \\
    CV-EN & 16K utt. & 27 & 1173 & 1125 & 169 \\
    CV-DE & 16K utt. & 27 & 1985 & 1906 & 304 \\
    CV-FR & 16K utt. & 26 & 600 & 549 & 225 \\
    CV-ES & 15.5K utt. & 26 & 122 & 135 & 15 \\
    \bottomrule
  \end{tabular}}
  \label{tab:test_sets}
\end{table}

\noindent \textbf{Dataset description} \quad Our experiments require keyword lists for biasing in addition to audio and transcripts. Publicly available test sets meeting this criterion are limited, and mostly in English. We evaluate our biasing approach on one private dataset (DefinedAI\footnote{Website: \url{https://www.defined.ai}}) covering banking, insurance and healthcare domains, and two public datasets: Earnings21\footnote{Audio split into 3-minute segments for decoding as in~\cite{drexler2022improving}.} (stock market) \cite{delrio2021earnings21} and CommonVoice~\cite{ardila2019common}; see Tab.~\ref{tab:test_sets}. DefinedAI provides manually annotated NEs. While Earnings21 includes two biasing lists from NER,\footnote{We use only the \textit{oracle} list from~\cite{drexler2022improving}.} CommonVoice lacks gold NEs, requiring us to generate our own bias lists.

\noindent \textbf{Biasing lists for CommonVoice} \quad We generate bias lists using HuggingFace BERT models~\cite{wolf2019huggingface,lhoest2021datasets} fine-tuned for NER per language.\footnote{Models: ES: \url{mrm8488/bert-spanish-cased-finetuned-ner}, EN: \url{dslim/bert-base-NER-uncased}, FR: \url{cmarkea/distilcamembert-base-ner}, DE: \url{fhswf/bert_de_ner}.} Extracted NEs are filtered to remove unigrams, reducing noise. See Tab.~\ref{tab:test_sets} for statistics. Experiments on CommonVoice assess (1) SF performance on languages other than English (EN) and (2) the effect of large biasing lists, e.g., the German (DE) subset has ~2k unique entities. Notably, DE and EN contain more unique NEs than French (FR) and Spanish (ES).

\subsection{Base ASR Model}
\label{subsec:training}

\noindent \textbf{Transformer-Transducer Training} \quad For all experiments, we utilize the stateless version of the Zipformer transducer model~\cite{stateless-predictor,zipformer}. 
For evaluation on the DefinedAI and Earnings21 test sets, we employ the pretrained Zipformer model trained on the Gigaspeech-XL dataset~\cite{chen2021gigaspeech}.\footnote{\raggedright Gigaspeech-XL: 10k hours of transcribed audio data, model: {\scriptsize \url{yfyeung/icefall-asr-gigaspeech-zipformer-2023-10-17}}}
This choice is justified because the Earnings21 dataset provides only test data, while for DefinedAI, we have access to limited training data (40 hours). 
Our choice of the model aligns with prior work on Earnings21, where the authors also used the Gigaspeech dataset for training~\cite{drexler2022improving}. This setup can be viewed as a domain adaptation scenario, given that both DefinedAI and Earnings21 consist of domain-specific data.

For experiments on CommonVoice, we train Zipformer models for each language using the corresponding training set. Training is performed from scratch using the latest Icefall Transducer recipe with its default hyperparameters. This includes the \textit{ScaledAdam} optimizer~\cite{kingma2014adam} and a learning rate scheduler with a 500-step warmup phase~\cite{vaswani2017attention}, followed by a decay phase determined by the total number of steps (7,500) and epochs (3.5)~\cite{zipformer}. The neural Transducer model is jointly optimized using an interpolated loss function, combining simple and pruned RNN-T loss~\cite{pruned-rnnt-loss, graves2012sequence} with CTC loss~\cite{ctc_loss} ($\lambda = 0.1$). The peak learning rate is set to $5.0 \times 10^{-2}$, and each model is trained for 30 epochs on a single RTX 3090 GPU. 

\begin{table}[t]
\caption{Biasing experimental results on DefinedAI and Earnings21 datasets. Results are reported in terms of WER and NE recognition accuracy (NE-a). Best performance appears in \textbf{bold} font, the second-best result appears \underline{underlined}. $^{\dagger}$The result is statistically significant with $p<0.07$. 
}
\label{tab:shallow_fusion_main}
\scriptsize
\centering
\resizebox{0.8\textwidth}{!}{
\begin{tabular}{cc cc cc}
    \toprule
    \multirow{3}{*}{\textbf{\shortstack{Decoding\\type}}}& \multirow{3}{*}{\textbf{\shortstack{Context\\source}}}& \multicolumn{2}{c}{\textbf{DefinedAI}} & \multicolumn{2}{c}{\textbf{Earnings21}} \\
    \cmidrule(lr){3-6}
    & & \textbf{WER} ($\downarrow$) & \textbf{NE-a} ($\uparrow$) & \textbf{WER} ($\downarrow$) & \textbf{NE-a} ($\uparrow$) \\
    
    \midrule
     \textbf{BeamS} & n/a &10.4 & 68.0 & 14.4 & 59.5 \\
     \multirow{2}{*}{\rotatebox[origin=c]{90}{\textbf{SF}}} & NNLM &\underline{10.2} & 69.3 & 14.9 & 58.5 \\
     & BPE-5gramLM&\underline{10.2} & 68.2 & 16.8 & 59.0 \\
    \midrule
    \midrule
    \multirow{3}{*}{\rotatebox[origin=c]{90}{\textbf{\shortstack{SF\,+\\AhoC}}}} & WL-3gramLM &\textbf{10.0} & 70.0 & \textbf{12.9} & 61.7 \\
     & KB &10.4 & \textbf{77.9$^{\dagger}$} & 16.7 & \textbf{63.5$^{\dagger}$} \\
     &  WL-3gramLM + KB&\textbf{10.0$^{\dagger}$} & \underline{73.3$^{\dagger}$} & \underline{13.1$^{\dagger}$} & \underline{62.8$^{\dagger}$} \\
    \bottomrule
    \end{tabular}
    }
\end{table}

\begin{table}[t]
    \centering
    \caption{Results of SF experiments on 4 languages (English, German, French, Spanish) of CommonVoice reported in terms of WER, recognition accuracy of NEs (NE-A) and OOV (OOV-A) words. Recent Whisper (small-S and medium-M) performance is given as a reference~\cite{whisper_model}. Best performance appears in \textbf{bold} font, while the second-best result appears \underline{underlined}. $^{\dagger}$The result is statistically significant with $p<0.07$.
    }
    \label{tab:sf_cg_commonvoice}
    \renewcommand{\arraystretch}{1.3}
    \resizebox{1.0\textwidth}{!}{
    \begin{tabular}{cccccccc}
    \toprule
    \textbf{Decoding type} & \textbf{Context source} & \textbf{WER}\,($\downarrow$) & \textbf{NE-A}\,($\uparrow$) & \textbf{OOV-A}\,($\uparrow$) & \textbf{WER}\,($\downarrow$) & \textbf{NE-A}\,($\uparrow$) & \textbf{OOV-A}\,($\uparrow$)\\
    \midrule
    
    & & \multicolumn{3}{l}{\textbf{CommonVoice-English}} & \multicolumn{3}{l}{\textbf{CommonVoice-German}} \\
    \multicolumn{2}{l}{\textit{Whisper-S (244M)/Whisper-M (769M)}} & 14.5/11.2 & - & - & 13.0/8.5 & - & - \\
    \cline{1-2}
    BeamS & n/a & 13.5 & 45.2 & 6.5 & 7.7 & 55.5 & 19.4 \\
    \multirow{2}{*}{SF}& NNLM & \underline{13.3} & 46.7 & 6.0 & \underline{7.6} & 55.5 & 18.8 \\
    & BPE-5gram-LM & \underline{13.3} & 45.4 & 6.6 & \underline{7.6} & 55.8 & 20.4 \\
    \cline{1-2}
    \multirow{3}{*}{\shortstack{SF\,+\\AhoC}} & WL-3gram-LM & \underline{13.3} & 46.9 & 5.9 & \underline{7.6} & 57.0 & 19.7 \\
    & Keyword boosting (KB) & 13.7 & \textbf{69.3$^{\dagger}$} & \textbf{36.0} & 7.7 & \textbf{79.0$^{\dagger}$} & \textbf{53.4} \\
    & WL-3gram-LM + KB & \textbf{13.2$^{\dagger}$} & \underline{59.3$^{\dagger}$} & \underline{24.1} & \textbf{7.4$^{\dagger}$} & \underline{70.8$^{\dagger}$} & \underline{39.7} \\

    \midrule
    \midrule
    & & \multicolumn{3}{l}{\textbf{CommonVoice-French}} & \multicolumn{3}{l}{\textbf{CommonVoice-Spanish}} \\
    \multicolumn{2}{l}{\textit{Whisper-S (244M)/Whisper-M (769M)}} & 22.7/16.0 & - & - & 10.3/6.9 & - & -  \\
    \cline{1-2}
    BeamS & n/a & 10.0 & 35.6 & 12.4 & 7.8 & 67.7 & 20.0 \\
    \multirow{2}{*}{SF}& NNLM & \textbf{9.8} & 36.1 & 11.5 & \underline{7.7} & 67.6 & 21.4 \\
    & BPE-5gram-LM & \textbf{9.8} & 35.8 & 12.9 & \textbf{7.6} & 66.9 & 23.1 \\
    \cline{1-2}
    \multirow{3}{*}{\shortstack{SF\,+\\AhoC}}& WL-3gram-LM & \textbf{9.8} & 37.3 & 13.2 & \textbf{7.6} & 66.9 & 18.8 \\
    & Keyword boosting (KB) & \underline{9.9} & \textbf{69.8$^{\dagger}$} & \textbf{53.9} & 7.8 & \textbf{86.8$^{\dagger}$} & \textbf{58.3} \\
    & WL-3gram-LM + KB & \textbf{9.8$^{\dagger}$} & \underline{53.6$^{\dagger}$} & \underline{33.0} & \textbf{7.6$^{\dagger}$} & \underline{80.2$^{\dagger}$} & \underline{35.7} \\

    \bottomrule
    \end{tabular}
    }
\end{table}


\begin{table}[t]
 \scriptsize
  \caption{Ablation of decoding speed (RTFX; higher better).}
  \label{tab:rtfx-oov-eval}
  \centering
    \resizebox{0.5\textwidth}{!}{
      \begin{tabular}{ ccc }
      \toprule
    \textbf{Decoding type} &\textbf{Context source}& \textbf{RTFX}($\uparrow$) \\
    \toprule
    \textbf{BeamS}&n/a & \textbf{120.7} \\
    \textbf{SF}&BPE-5gramLM & 77.8 \\
    \midrule
    \midrule
    \multirow{3}{*}{\rotatebox[origin=c]{90}{\textbf{\shortstack{SF\,+\\AhoC}}}}&WL-3gramLM & 111.1 \\
    &KB & 113.5 \\
    &WL-3gramLM + KB & \underline{117.3} \\
    \bottomrule
\end{tabular}
}
\end{table}

\subsection{Experiments and Evaluation}
For all experiments, we utilize their respective Zipformer-Transducer models (70M), with variations occurring solely in the decoding strategies. We benchmark our proposed approach against established methods for integrating external context during decoding.  

\noindent \textbf{Baselines} \quad We implement three baseline methods, ranging from standard beam search to advanced SF (\textbf{SF}) techniques incorporating neural network language models (NNLMs) for contextual adaptation. $\bullet$ \textbf{BeamS}: Standard beam search decoding without any external contextual information (i.e., no SF). $\bullet$ \textbf{NNLM}: A conventional SF approach integrating a Transformer-based NNLM. $\bullet$ \textbf{BPE-5gramLM}: A SF setup using a 5-gram LM trained at the BPE level.

\noindent \textbf{Context Biasing with Aho–Corasick} (\textbf{SF+AhoC}) \quad We evaluate three variations of our proposed contextual adaptation method, leveraging the AC algorithm for efficient integration of external context. $\bullet$ \textbf{WL-3gramLM}: SF with a 3-gram word-level LM, integrated via an AC-based trie for fast lookup and retrieval. $\bullet$ \textbf{KB}: Context biasing using an AC-based trie for keyword matching, without an explicit LM. $\bullet$ \textbf{WL-3gramLM+KB}: a unified fusion approach that integrates a 3-gram word-level LM and a KB list through a single AC-based decoding trie pass.

\noindent \textbf{LMs} \quad  
For training the NNLM used in our baselines, we train Transformer-based~\cite{vaswani2017attention} LMs. We use GigaSpeech-XL for \mbox{DefinedAI} and Earnings21, while for CommonVoice, we use the corresponding language-specific training sets. Each Transformer LM is trained for 10 epochs and consists of approximately 38M parameters.\footnote{Transformer-LM recipe: \url{https://github.com/k2-fsa/icefall/tree/master/icefall/transformer_lm}} Regarding the BPE-5gramLM, we train a 5-gram BPE-based LM using the SRILM~\cite{stolcke2002srilm} toolkit.  

\noindent Similarly, for the word-level LMs required for the \textbf{SF+AhoC} experiments, we train 3-gram word-level LMs with SRILM. To construct these models, we use text data from the corresponding training sets for all test sets except Earnings21. For Earnings21, we instead use transcriptions from Earnings22~\cite{del2022earnings}, a dataset from the same domain, ensuring domain consistency.

\noindent \textbf{Evaluation} \quad In addition to the word error rate (WER) metric, we evaluate the accuracy only on NEs (\textit{NE-a}). To calculate the NE metrics, only strings containing NEs in the references are taken into account. Accuracy is calculated in a binary manner: ``yes'' -- when the NE is completely recognized correctly, ``no'' -- when at least one error occurs within the NE. For NE-A results with KB and WL-3gramLM+KB experiments and the WER results with WL-3gramLM+KB experiments, we also report statistical significance tests against the baseline (BeamS). For WER evaluation on Earnings21, we use the \textit{fstalign tool}\footnote{Provided by authors of~\cite{delrio2021earnings21} as the dataset references are in a special NLP-format: \url{https://github.com/revdotcom/fstalign}}  and for NE accuracy we used only ``PERSON'' and ``ORG'' categories. To evaluate OOV performance, we measured the recognition accuracy of OOV-NEs for German and French, as their datasets have the highest numbers of OOV-NEs (see the last column in Tab.~\ref{tab:test_sets}).

As fast and flexible context integration is critical in many practical scenarios, we include an estimate of decoding time using the \textit{inverse real-time factor (RTFX)}, which is the ratio between the length of the processed audio and the decoding time. RTFX is measured on the DefinedAI test set with one RTX 3090 GPU.

\section{Results}
\label{sec:results}

\textbf{The impact of word-level n-grams} \quad To distinguish between out-of-domain and in-domain performance, we present results on DefinedAI and Earnings21 (Tab.~\ref{tab:shallow_fusion_main}) separately from CommonVoice (Tab.~\ref{tab:sf_cg_commonvoice}) experiments. 
For both setups, the results of fusion n-gram LM into a context trie (i.e., WL-3gramLM) lead to relative WER reduction w.r.t.\ beam search alone: 3.8\% for DefinedAI, 10.4\% for Earnings21 (see Tab.~\ref{tab:shallow_fusion_main}), 1.5\%, 1.3\%, 2\%, and 2.6\% for EN, DE, FR, and ES from CommonVoice respectively (see Tab.~\ref{tab:sf_cg_commonvoice}).
Moreover, for the out-of-domain sets, it improves the performance compared to the fusion with Transformer-LM (NNLM): i.e., from 10.2 to 10.0 for DefinedAI and from 14.9 to 12.9 for Earnings21. In addition to improved accuracy, training n-gram LMs is fast and easy and can be done even with a small corpus (e.g., 40~hours in the case of DefinedAI), making it suitable for low-resource scenarios.

\noindent \textbf{Trade-off between NE-accuracy and WER} \quad The biggest improvement on NEs is always achieved with the KB setup: 14.6\% for DefinedAI and 6.7\% for Earnings21 of relative improvement w.r.t.\ BeamS baseline (Tab.~\ref{tab:shallow_fusion_main}). However, the overall WER does not improve or even degrades, e.g., to 16.7 WER in the case of Earnings21. The overall WER is improved when KB is combined with word-level n-gram LM (WL-3gramLM + KB): relative improvement w.r.t. KB alone is by 3.8\%, 21.6\%, 3.6\%, 3.9\%, 1\%, and 2.6\% for DefinedAI, Earnings21, and CommonVoice EN, DE, FR and ES, respectively. It is important to note that the datasets vary in the number of NEs and the proportion of utterances containing them (see Tab.~\ref{tab:test_sets}). This variation explains the smaller impact of fusion on WER for the FR and ES sets, which contain the fewest NEs. Nevertheless, even in these cases, our approach maintains WER performance without degradation.


\noindent \textbf{RTFX} \quad The RTFX results in Tab.~\ref{tab:rtfx-oov-eval} show that decoding with \textit{WL-3gramLm + KB} is only a bit slower compared to BeamS approach alone and thus it can be used on-the-fly:
RTFX of the \textit{WL-3gramLm + KB} method is 6.0\% lower than the \textit{baseline} (BeamS) and there is no degradation w.r.t \textit{KB}.
Finally, while the WER performance of BPE-5gramLM and WL-3gramLM is generally comparable across datasets, a key distinction lies in their computational overhead. WL-3gramLM is 31\% faster than BPE-5gramLM, with an RTFX of 77.8 compared to 111.1, offering a significant advantage for production-level ASR solutions.

\noindent \textbf{OOV performance} \quad The incorporation of word-level statistics via WL-3gramLM does not adversely affect OOV recognition at the subword level (as indicated by the \textbf{OOV-A} column in Table~\ref{tab:sf_cg_commonvoice}). Despite the KB yielding optimal performance, it may not be the most suitable option if overall WER is a concern. Similar to named entities, the WL-3gramLM+KB configuration strikes an effective balance, preserving overall ASR performance while enhancing OOV accuracy by up to 51\% for the German dataset. These outcomes substantiate that, even with the integration of word-level data into the rescoring process, the benefits of fusion extend to unseen words, given that the biasing occurs at the subword level.

\section{Conclusion}
\label{sec:conclusion}
We present an efficient approach for integrating word-level n-gram LMs with a Transformer Transducer ASR model using SF and a trie based on the Aho-Corasick algorithm. The method enables faster decoding than standard SF maintaining competitive WER and real-time factor (RTF) efficiency. Additionally, we show that combining n-gram LMs with KB in a unified context graph improves overall WER and NE accuracy, including for OOV words. Experiments across four languages and three datasets confirm the effectiveness of our approach in both in-domain and out-of-domain scenarios.

\bibliographystyle{splncs04}
\bibliography{paper}

\end{document}